\documentclass[journal]{IEEEtran}

\ifCLASSINFOpdf

\else

\fi
\usepackage{geometry}
\usepackage[T1]{fontenc}
\usepackage{ulem}
\usepackage{amsmath,amssymb,amsfonts}
\usepackage{algorithm}
\usepackage{algpseudocode}
\usepackage{graphicx}
\usepackage{textcomp}
\usepackage{xcolor}
\usepackage{mathrsfs}
\usepackage{bm}
\usepackage{threeparttable}
\usepackage{float}
\usepackage{stfloats}
\usepackage{array}
\usepackage{marvosym}
\usepackage {setspace}
\usepackage[T1]{fontenc}
\usepackage[utf8]{inputenc}
\usepackage{babel}
\usepackage{booktabs}
\usepackage{threeparttable}
\usepackage{multirow}

\makeatletter

\newcommand{\Rmnum}[1]{\expandafter\@slowromancap\romannumeral #1@}
\makeatother

\begin{document}

\title{From Point to Space: 3D Moving Human Pose Estimation Using Commodity WiFi}
\author{Yiming Wang, Lingchao Guo, Zhaoming Lu, 
	    Xiangming Wen, Shuang Zhou, and Wanyu Meng

\thanks{This work was supported by National Natural
	Science Foundation of China under Grant 61671073. \itshape{(Corresponding author: Zhaoming Lu.)}}
\thanks{The authors are with the School of Information and Communication Engineering,
	also with the Beijing Key Laboratory of Network System Architecture
	and Convergence, and also with the Beijing Laboratory of Advanced Information
	Networks, Beijing University of Posts and Telecommunications, Beijing 100876,
	China (e-mail: lzy0372@bupt.edu.cn).}}

\maketitle

\normalem
\begin{abstract}
In this paper, we present Wi-Mose, the first 3D moving human pose estimation system using commodity WiFi. Previous WiFi-based works have achieved 2D and 3D pose estimation. These solutions either capture poses from one perspective or construct poses of people who are at a fixed point, preventing their wide adoption in daily scenarios. To reconstruct 3D poses of people who \emph{move throughout the space rather than a fixed point}, we fuse the amplitude and phase into \emph{Channel State Information (CSI) images} which can provide both pose and position information. Besides, we design a neural network to extract features that are only associated with poses from CSI images and then convert the features into key-point coordinates. Experimental results show that Wi-Mose can localize key-point with 29.7mm and 37.8mm Procrustes analysis Mean Per Joint Position Error (P-MPJPE) in the Line of Sight (LoS) and Non-Line of Sight (NLoS) scenarios, respectively, achieving higher performance than the state-of-the-art method. The results indicate that Wi-Mose can capture high-precision 3D human poses throughout the space.

\end{abstract}
\vspace{-0.1cm}
\begin{IEEEkeywords}
WiFi sensing, 3D human pose estimation, CSI images, neural network design.
\end{IEEEkeywords}
\vspace{-0.2cm}

\IEEEpeerreviewmaketitle

\section{Introduction}

\IEEEPARstart{I}{n} recent years, thanks to the increasingly ubiquitous deployment of WiFi infrastructure and the open-source software \cite{Tool}, human sensing based on WiFi Channel State Information (CSI) has gained significant attention \cite{track} \cite{wirol} \cite{fall}. To achieve more fine-grained human sensing, WiFi-based human pose estimation has been a  focus of research recently. Pioneering works \cite{rf} \cite{rf3D} estimate human poses using radars operating at the WiFi frequency, i.e., 5.46-7.24$GHz$. \par
Further, for easy deployment in real world and cost savings, \cite{letters} \cite{mobicom} use commodity WiFi devices to obtain fine-grained human poses. \cite{letters} enables commodity WiFi devices to capture 2D human skeleton images. But it can merely obtain human poses from one perspective, and performs unsatisfactorily in some special perspectives restricted by annotations.\par 
Different from \cite{letters}, \cite{mobicom} presents WiPose, the only 3D human pose estimation system using commodity WiFi. It utilizes the amplitude component of CSI and achieves high performance in the experimental environment. However, during the experiment, the subjects are required to perform movements at a fixed point. Moreover, it consists of 9 distributed antennas. Therefore, WiPose is limited to some specific applications and not convenient enough for daily use, such as smart home, health monitoring, etc.\par
In summary, due to the change of the target position and the difficulty of data collection, WiFi-based human pose estimation still has some limitations for wide adoption in daily life.\par
To address the above limitations, we propose Wi-Mose, the first system which can capture fine-grained 3D moving human poses with commodity WiFi devices in both Line of Sight (LoS) and Non-Line of Sight (NLoS) scenarios. To reconstruct 3D moving human poses, we mainly design the system from two aspects: data processing and network design. \par
For one thing, we convert the processed amplitude and phase, which contain the pose and position information, respectively, into a sensitive CSI tensor, called \emph{CSI images}. And then, we feed CSI images into the network rather than only amplitude or phase information. For another, we design a deep feature extraction network to extract pose related features from the amplitude channel and weaken the influence of position changes by leveraging the information in the phase channel. Specifically, we use the position information contained in the phase channel as prior knowledge to add constraints to the attitude estimation. We also design a pose regress network to convert the features into key-point coordinates. Therefore, the subject can \emph{move continuously and freely without space constraints}. The main contributions of our work are listed as follows:\par
1. We propose a method to convert the raw CSI data into CSI images so that the neural network can extract features which contain more pose information but less position component. \par
2. We design a neural network which is suitable to extract moving human pose features from CSI images and convert WiFi signals into 3D human poses.\par
3. We build a 3D human pose estimation prototype system for experiment and evaluation. Results show that the system can estimate 3D human poses with 29.7mm (37.8mm) Procrustes analysis Mean Per Joint Position Error (P-MPJPE) in the LoS (NLoS) scenarios, achieving 21\% (10\%) improvements in accuracy over the state-of-the-art method. \par
4. Because of the usage of CSI images and specialized network, Wi-Mose utilizes only 6 antennas to capture information, which is lightweight and low-cost compared with the state-of-the-art method \cite{mobicom}.\par
The rest of this paper is organized as follows. Section II is the system overview. Section III discusses data collection and processing. Section IV introduces the neural network. Section V is the baseline. Section VI describes experiments and performance followed by a conclusion in Section VII.


\begin{figure*}[!t]
	\vspace{-0.9em}
	\setlength{\abovecaptionskip}{-0.1cm}
	\setlength{\belowcaptionskip}{0.cm}
	\centerline{\includegraphics[scale=0.58]{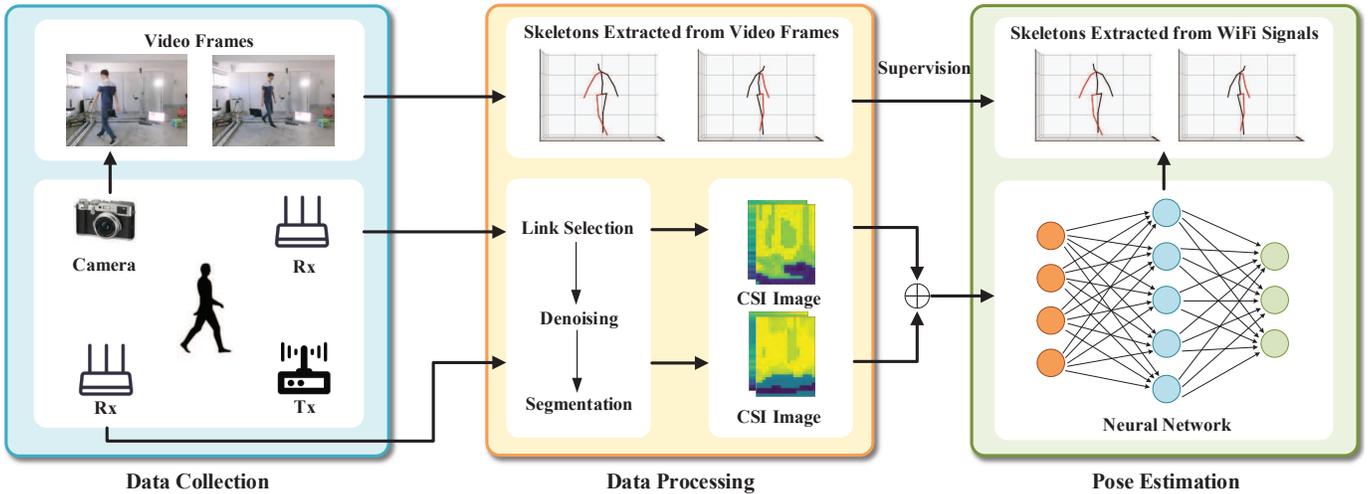}}
	\caption{The system overview. The upper pipeline provides ground truth for supervision training, while the bottom pipeline learns to extract human poses using only WiFi signals from commodity devices.}
	\label{system}
	\vspace{-1.8em}
\end{figure*}
\section{System Overview}
\vspace{-0.1cm}
The system consists of three parts: data collection, data processing, and pose estimation, as shown in Fig.~\ref{system}. The data collection part contains two receivers, a transmitter, and a monocular camera, which are used to collect synchronous CSI and video frames. The data processing part converts the raw CSI data into CSI images and transforms the video frames to human key-point coordinates which are used for supervised learning. The pose estimation part extracts features from CSI images and converts the features into key-point coordinates which are utilized to reconstruct 3D human pose skeletons.
\vspace{-0.3cm}

\section{Data Collection and Processing}
\vspace{-0.2cm}
In this paper, to reconstruct 3D moving human pose skeletons using commodity WiFi devices, we need to collect synchronous raw CSI and video frames and then process these data so that they can be fed into the neural network.\\
\vspace{-1cm}
\subsection{Data Collection}
\vspace{-0.1cm}
According to Fresnel zone model \cite{fresnel}, to capture human poses in the whole space, we need at least 2 pairs of transceivers. Hence in this paper, we utilize 3 commodity WiFi devices (one transmitter and two receivers) to capture the raw CSI data which contains human pose information. To collect more information, we set 1 transmitting antenna at the transmitter, 3 receiving antennas at each receiver. In order to improve resolution and capture pose-related features efficiently, the 2 pairs of transceivers are mutually perpendicular and the transmitter is placed at the intersection. During collecting CSI data, we collect synchronous video frames using a monocular camera, which are utilized to extract 3D key-point coordinates as the ground truth to train the proposed neural network.
\vspace{-0.5cm}
\subsection{Data Processing}
\vspace{-0.1cm}
\subsubsection{Link Selection}We observe that there is always an antenna which receives CSI with a larger variance value than others. It means this antenna has larger dynamic responses. So we choose it as the reference and make use of its amplitude information. 

\subsubsection{Denoising}
In reality, there are multiple paths between a pair of transceiver. In an ideal state, the response of the wireless channel at time $t$ and frequency $f$ can be expressed as:

\begin{footnotesize}
	\vspace{-0.1cm}
	\begin{equation}
		H(f,t)=\sum_{i=1}^{N}\alpha _{i}(t)e^{-j2\pi f\tau_{i}(t)}   
	\end{equation}
\vspace{-0.3cm}
\end{footnotesize}\par

\noindent where $N$ is the number of multipath, $\alpha _{i}(t)$ and $\tau_{i}(t)$ are the complex attenuation and time of flight for the $i$-th path, respectively.\par
According to whether the length of the path changes, CSI can be divided into two parts, the static path and the dynamic path components, which can be expressed as:

\begin{footnotesize}
	\vspace{-0.1cm}
	\begin{equation}
		H(f,t)=H_{s}(f,t)+\sum_{i\in P_{d}}\alpha _{i}(t)e^{-j2\pi f\tau_{i}(t)}   
	\end{equation}
	\vspace{-0.3cm}
\end{footnotesize}\par

\noindent where $H_{s}(f,t)$ is the sum of responses of all static paths including LoS and other static reflection paths, $P_{d}$ is the collection of dynamic paths which are not constant over time. Our purpose is to extract the dynamic path component.\par
We cannot directly utilize raw CSI to capture human poses. Because compared with the signals of LoS and other static paths in raw CSI measurements, pose-related signals are too week and easily influenced by unpredictable interference. To improve the accuracy of pose estimation, we eliminate interference and extract the dynamic paths corresponding to humans.\par
For more denoising details, please refer to the previous work of our team in \cite{letters}.
\subsubsection{Segmentation}In order to capture continuous human poses, we segment the processed CSI according to the synchronous video frames and reconstruct CSI images, as shown in Fig.~\ref{system}. The CSI images contain an amplitude channel and a phase channel. The amplitude can reflect people's movements, while the phase is more used to obtain changes in position. To extract pose information and weaken the influence of position changes, we combine them to provide both pose and position information for the neural network. \par

\vspace{-0.1cm}
\begin{figure}[!t]
	\vspace{-0.5cm}
	\setlength{\abovecaptionskip}{-0.1cm}
	\setlength{\belowcaptionskip}{0.cm}
	\centerline{\includegraphics[scale=0.8]{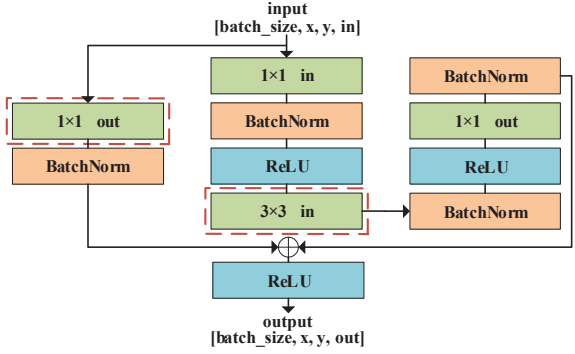}}
	\caption{The structure of a residual block. The 1$\times$1 and 3$\times$3 refer to the kernel size. In and out refer to the out channel of the layer.}
	\label{block}
	\vspace{-1em}
\end{figure}

\section{Neural Network Design}
\vspace{-0.1cm}
In this paper, we design a neural network to extract features and convert them into key-point coordinates. In order to reduce the influence of position changes on pose estimation, different from \cite{mobicom}, we choose to directly regress the key-points, similar to the method in computer vision.
\vspace{-0.3cm}

\subsection{Data and Annotations}
\vspace{-0.1cm}
To accurately and intuitively associate CSI data with human poses, we use a camera synchronized with a receiver to capture video frames. Then, we apply AlphaPose \cite{alpha} and VideoPose3D \cite{videopose} to get the 3D key-point coordinates from the video frames. Since our goal is to reconstruct 3D moving human pose skeletons, we choose key-point coordinates as the annotation rather than the whole skeleton. Because the limbs are rigid, locating key-points can be more accurate and prevent overfitting.

\vspace{-0.4cm}
\subsection{Network Framework}
\vspace{-0.1cm}
The design of our neural network must consider the time correlation of human poses and the spatial position of the human body. In addition, the spatial resolution of WiFi signals is low, which makes it difficult to capture complete human poses from just a single CSI sample. To solve these problems, we make the network learn to aggregate information from multiple CSI samples instead of taking a single CSI sample as input. \par
We design a neural network to convert CSI data into 3D human key-point coordinates, including a feature extraction network and a key-point regression network. Because the input CSI images contain a phase channel which increases the amount of data and introduces linear constraints, the feature extraction network should be able to extract sufficiently deep features.\par 

\begin{figure*}[!t]
	\vspace{-1em}
	\setlength{\abovecaptionskip}{-0.1cm}
	\setlength{\belowcaptionskip}{-0.0cm}
	\centerline{\includegraphics[scale=0.356]{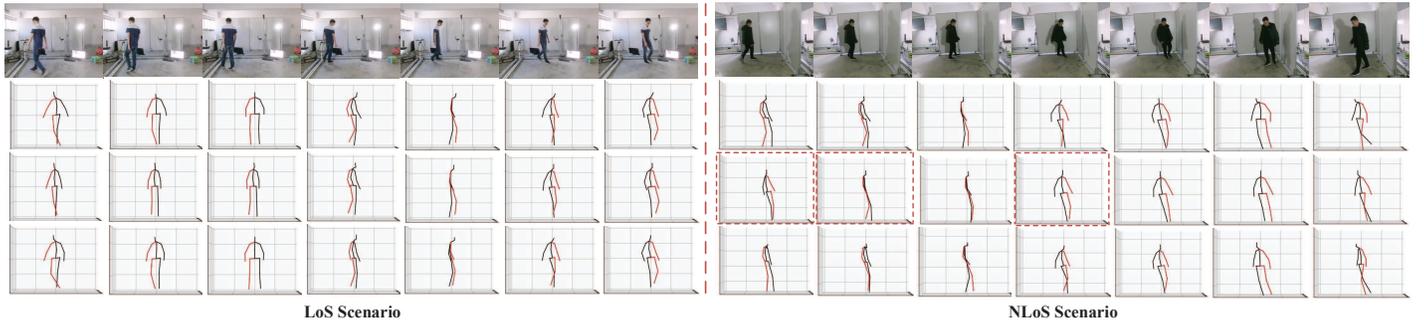}}
	\caption{Test examples which show the constructed skeletons of a person in the LoS and NLoS scenarios. The first line: Video frames captured by the camera, and presented here for visual reference. The second line: Human poses extracted from video frames, and presented here as ground truth. The third line: The human poses captured by WiPose. The fourth line: The human poses captured by Wi-Mose. The poses in red boxes have obvious errors.}
	\label{show}
	\vspace{-1.5em}
\end{figure*}

Therefore, in the feature extraction network, we use a residual network, which contains 13 residual blocks as shown in Fig.~\ref{block}, to extract features related to human poses. Because of its special structure, the residual network can avoid gradient explosion and disappearance caused by the deepening of the network. In the key-point regression network, we utilize two fully connected layers to integrate feature information, and finally convert these features into key-point coordinates. The details of the network are shown in Table~\ref{network}.\par

Take a set of synchronized CSI data and key-point coordinates $(C_{1},C_{2},K)$ as an example, where $(C_{1},C_{2})$ denotes the CSI images from two pairs of transceivers and $K$ denotes the corresponding key-point coordinates from video frames.\par
In the training stage, we feed $(C_{1},C_{2})$ into the proposed neural network and get the predicted 3D human key-point coordinates $P$. Then we use $K$ as an annotation and compare $P$ with it to optimize the entire network. \par
We define the training process as minimizing the average Euclidean distance error between predicted joints and the ground truth, so we first define the position loss $L_{P}$ as the $L_{2}$ norm between the predicted joints and the ground truth:

\begin{footnotesize}
	\vspace{-0.1cm}
	\begin{equation}	
		L_{P} =\frac{1}{T}\sum_{t=1}^{T}\frac{1}{N}\sum_{i=1}^{N}\left \| \tilde{p}_{t}^{i}-p_{t}^{i}  \right \|_{2} ,
	\end{equation}
\vspace{-0.3cm}
\end{footnotesize}

\noindent where $\tilde{p}_{t}^{i}$ and $p_{t}^{i}$ are the predicted and real coordinate of joint $i$ in time slot $t$, $N$ is the joint number in the model we use, and $T$ means the number of data samples.\par

\begin{table}[t]
	\centering
	\setlength{\abovecaptionskip}{-0.1cm}
	\vspace{-0.5cm}
	\setlength{\belowcaptionskip}{0.cm}
	\caption{The Neural Network Implementation}
	\label{network}
	\renewcommand{\arraystretch}{1.3}
	\renewcommand\tabcolsep{16.0pt}
	\fontsize{7}{7}\selectfont
		\begin{tabular}{c|c|c|c}
			\toprule
			Network &Input Size & Output Size  & Stride \\
			\hline
			BLOCK1& 30$\times$20$\times$4& 30$\times$20$\times$4& 1$\times$1 \\
			BLOCK2 &30$\times$20$\times$4& 30$\times$20$\times$8& 1$\times$1 \\
			BLOCK3 &30$\times$20$\times$8& 15$\times$10$\times$8& 2$\times$2 \\
			BLOCK4 &15$\times$10$\times$8& 15$\times$10$\times$16& 1$\times$1 \\ 
			BLOCK5 &15$\times$10$\times$16& 8$\times$5$\times$16& 2$\times$2 \\ 
			BLOCK6 &8$\times$5$\times$16& 8$\times$5$\times$64& 1$\times$1 \\ 
			BLOCK7 &8$\times$5$\times$64& 4$\times$3$\times$64& 2$\times$2 \\ 
			BLOCK8 &4$\times$3$\times$64& 4$\times$3$\times$256& 1$\times$1 \\
			BLOCK9 &4$\times$3$\times$256& 2$\times$2$\times$256& 2$\times$2 \\ 
			BLOCK10 &2$\times$2$\times$256& 2$\times$2$\times$1024& 1$\times$1 \\ 
			BLOCK11 &2$\times$2$\times$1024& 1$\times$1$\times$1024& 2$\times$2 \\
			BLOCK12 &1$\times$1$\times$1024& 1$\times$1$\times$2048& 1$\times$1 \\
			BLOCK13 &1$\times$1$\times$2048& 1$\times$1$\times$2048& 1$\times$1 \\
			FC1 &1$\times$1$\times$2048& 1$\times$512& - \\ 
			FC2 &1$\times$512& 1$\times$51& - \\ 
			\bottomrule	
	\end{tabular}
	\begin{tablenotes}
		\scriptsize
		
		\item[1] 1. Stride just applies to the two layers inside the red box in Fig.~\ref{block}.
		\item[2] 2. Other layers' stride is always $1\times1$.
	\end{tablenotes}
	
	\vspace{-3em}
\end{table}

Since our network regresses the key-points directly, we introduce Huber Loss in the loss function. Huber Loss is a parameterized loss function for regression problems. It can enhance $L_{P}$ and reduce the interference of outliers. The Huber Loss $L_{H} $ in our loss function is expressed as:

\begin{footnotesize}
	\vspace{-0.1cm}
	\begin{equation}
		L_{H} =\frac{1}{T}\sum_{t=1}^{T}\frac{1}{N}\sum_{i=1}^{N}\left \| \tilde{p}_{t}^{i}-p_{t}^{i}  \right \|_{H} ,
	\end{equation}
	\vspace{-0.1cm}
\end{footnotesize}

\noindent where $\left\|\cdot\right \|_{H}$ means the Huber norm. It is defined as:
\vspace{-0.1cm}
\begin{footnotesize}
	\begin{equation}
		\left\|x\right \|_{H}=\frac{1}{n}\sum_{i=1}^{n}huber(x_{i}),
	\end{equation}
\vspace{-0.1cm}
\end{footnotesize}\par

\noindent where:
	\vspace{-0.1cm}
\begin{footnotesize}
	\begin{equation}
		huber(x_{i})=
		\begin{cases}
			\quad 0.5x_{i}^{2}  \quad \quad\quad \!\!{\rm if}\;x_{i}<\delta \\
			\quad|x_{i}|-0.5 \quad {\rm otherwise.}
		\end{cases}
	\end{equation}
\end{footnotesize}\par
	\vspace{-0.1cm}

The $\delta$ is a parameter of the Huber Loss and set to 0.75 in our experiment.\par
Finally, the loss function is defined as:

\begin{footnotesize}
	\begin{equation}
		L=L_{P}+L_{H}
	\end{equation}

\end{footnotesize}\par
	\vspace{-0.1cm}
We use the Adam \cite{Adam} optimizer to optimize the loss function in our network.

\vspace{-0.5cm}
\subsection{Network Settings and Training}
We collect CSI data at 150Hz and video frames at 30Hz, which means every 5 CSI samples in each receiver are synchronized with one video frame by timestamps. Consider the continuity of movement and the efficiency of training, we use 5 strictly corresponding CSI samples and the preceding 15 samples to correspond to one video frame. Therefore, the final result is 20 unique CSI samples corresponding to one video frame, that is, the final generated action is 7.5 Hz. \par

The structure of the neural network is shown in Table~\ref{network}. The feature extraction part consists of 13 residual blocks shown in Fig.~\ref{block}. We add a batch normalization layer after each convolution. In order to add non-linearity to the model, we use Rectified Linear Unit (ReLU) activation functions after each batch normalization layer. To improve training efficiency, we set the stride of the two layers inside the red box in Fig.~\ref{block} to 2$\times$2 in the 3rd, 5th, 7th, 9th, and 11th residual blocks, and all other strides to 1$\times$1. In addition, in these layers, we perform a convolution operation (kernel size is 1$\times$1, stride is 2$\times$2) on the input to make it match the size of the output. After the residual network, two fully connected layers convert high-dimensional features into the key-point coordinates. \par
The network is implemented with TensorFlow \cite{tensorflow}. The result is trained for 6 epochs using the Adam \cite{Adam} optimizer with 0.0001 learning rate, 4 batch size. Moreover, we introduce learning rate decay method that the learning rate is multiplied by 0.9 after every epoch.

\vspace{-0.4cm}

\section{Baseline}
\vspace{-0.1cm}
WiPose proposed in \cite{mobicom} is currently the state-of-the-art 3D human pose estimation system based on commodity WiFi. In this paper, we apply the deep learning model proposed in WiPose as our baseline. The model contains four-layer Convolutional Neural Network (CNN) and three-layer Long Short Term Memory network (LSTM) on top of the CNN. The system applies the output of LSTM to the initial skeleton model to obtain the current poses according to forward kinematics. In \cite{mobicom}, the author proposes two input data formats. Because our system uses 2D CSI data, we only use 2D CSI data to train the model in \cite{mobicom}.  \par
Note that although our work is carried out on the basis of \cite{letters}, we do not choose the model in \cite{letters} as our baseline. Because the model estimates 2D human skeleton images while our goal is to estimate 3D human poses.
\vspace{-0.3cm}
\section{Experiment and Evaluation}
\vspace{-0.2cm}
\subsection{Setup}
Our experimental site is a 7m$\times$8m basement. We use 3 transceivers work in the 5GHz frequency with 20MHz bandwidth and we install CSI-Tool \cite{Tool} on all the transceivers. The two receivers are synchronized using the Network Time Protocol (NTP). And we connect a monocular camera to a receiver to record video frames. Throughout the experiment, other wireless signals are still operated.

\vspace{-0.4cm}

\subsection{Dataset}
\vspace{-0.1cm}
The dataset contains 10 hours of data for 5 people with different heights, weights, clothing, and genders, which means that we finally collected 5,400,000 CSI samples at each receiver. We ask each volunteer to perform continuous poses in the room. Meanwhile, the camera will record their poses. Then we obtain the joints synchronized with CSI samples through automatic annotations for joints (AlphaPose and VideoPose3D). For the data of the four persons who are chosen as the training subjects, 75\% is used as the training set to train the network and the remaining 25\% is used as the test set to test the model. The data of the last person is used for testing the generalization of our model.
\vspace{-0.2cm}
\begin{table*}[!htbp]
	\setlength{\abovecaptionskip}{-0.1cm}

	\renewcommand{\arraystretch}{1.1}
	\renewcommand\tabcolsep{2.5pt}
	\caption{P-MPJPE(unit:$mm$) for The Basic Scenario}
	\label{basic}
	\fontsize{8}{10}\selectfont  
	\centering
	
	\begin{tabular}{c|c|c|c|c|c|c|c|c|c|c|c|c|c|c|c|c|c|c}
		\toprule
		Joints & MidHip & LHip  & LKnee  & LAnkle & RHip & RKnee & RAnkle & Back & Neck & Nose  & Head & LShoulder & LElbow & LWrist & RShoulder & RElbow & RWrist &  Overall\\ \hline
		WiPose &17 & 30  & 36  & 57  & 34  & 40  & 60  &31 & 13  & 25  & 38 & 26 & 45 & 53 & 24 & 43 & 67 & 37.6\\ 
		 Wi-Mose &16 & 22  & 32  & 49 & 23 & 31 & 50 &18 & 14 & 24 & 36 &19 & 31 & 46 &18 &24 &52&29.7\\ 
		\bottomrule
	\end{tabular}
	
	\vspace{-2em}
\end{table*}

\vspace{-0.3cm}

\subsection{Performance}
\vspace{-0.1cm}
To measure the performance, we introduce P-MPJPE which performs Procrutes analysis before calculates MPJPE. We observe that compared with the ground truth, some constructed 3D human pose coordinates have slight and global offset. The reason is that we directly regress key-point coordinates, which introduces an error independent of poses. Since P-MPJPE is more suitable for moving human pose estimation, we utilize it to weaken the effect of the attitude-independent error.
\subsubsection{Basic Scenario}We first evaluate the performance in the basic scenario which is the LoS scenario. The left part of Fig.~\ref{show} shows a test example of the constructed skeletons of a person who continuously walks in different poses and directions.  \par
Table~\ref{basic} reports the P-MPJPE for the basic scenario. We not only calculate the errors for each joint but also measure the overall performance by averaging them. The results show that the proposed system performs much better than the baseline. The overall P-MPJPE of our system is 29.7mm, while that of the baseline is 37.6mm. Note that the error of the root joint in WiPose, which is set to Neck in our model, is smaller than that in Wi-Mose because of the introduction of forward kinematics. The positioning accuracy of other points of Wi-Mose is higher than that of the baseline. For both baseline and the presented system, it is more difficult to locate the joints farther from the trunk, since the reflected signals from these parts are weaker than those from closer parts. Another reason is that these parts of human body have smaller reflection areas and always have a larger moving range in our dataset, which makes it much harder to locate.\par
\begin{figure}[t]
	\setlength{\abovecaptionskip}{0.0cm}
	\setlength{\belowcaptionskip}{0.0cm}
	\centerline{\includegraphics[scale=0.32]{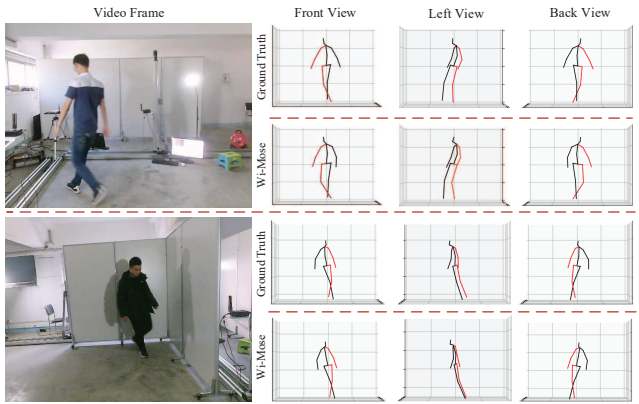}}
	\caption{The examples of estimated poses in multiple perspectives for the LoS (upper) and NLoS (bottom) scenarios.}
	\label{multi}
	\vspace{-0.8em}
\end{figure}

 
The upper part of Fig.~\ref{multi} shows a test example in different perspectives. Compared with 2D pose estimation, we can show the poses in all perspectives, even if the limbs are obscured. The results show that Wi-Mose is more suitable to capture 3D human pose throughout the space.

\subsubsection{Occluded Scenario}
In order to verify the performance of Wi-Mose in the occluded environment (the NLoS scenario), we add a wooden screen between the subject and the receiver. The distribution of training and test data is the same as the basic environment. A test example is shown in the right part of Fig.~\ref{show}. The overall P-MPJPE result of Wi-Mose is 37.8mm, while that of the baseline is 42.0mm, indicating that Wi-Mose outperforms the baseline. Thanks to the penetration of WiFi, we can still receive signals even if there are obstructions. Because the signals will attenuate when passing through the wooden screen, the final error will be larger than the basic scenario. In the NLoS scenario, some poses estimated by the baseline are obviously wrong. The reason is that the baseline is more susceptible to position changes than Wi-Mose, especially when there is less information. \par
The bottom part of Fig.~\ref{multi} shows a test example in different perspectives. The results indicate that Wi-Mose can capture high-precision 3D moving human poses in the occluded scenario.

\subsubsection{Cross-subject Scenario}
Wi-Mose performs well in cross-subject scenario. We collect 5 people's data in both the basic and occluded scenarios. When we train the network in each scenario, we feed the data of the first four people into the network, and then, use the last person's data for testing. It should be noted that there are obvious differences in height and weight of the five people. The result is in Table~\ref{cross}. As we can see, the performance of the proposed Wi-Mose framework is slightly better than the baseline. Compared with the basic and occluded scenarios, the performance of our system is degraded a little larger than that of the baseline. The reason is that we use the network loss to constrain the pose estimation, which is a weak constraint. While the baseline applies the preset skeleton model and the network loss to constrain the poses, which is a strong constraint. This makes our model more flexible in dealing with full-space scenarios, but less satisfactory in cross-subject pose estimation.


\section{Conclusion}

In this paper, we present Wi-Mose, the first high-precision 3D moving human pose estimation system using commodity WiFi devices. We construct CSI images which contain both pose and position information so that the neural network can extract features which is related to poses but independent of position. Moreover, we design a neural network to extract features from CSI images and convert them into key-point coordinates. The experiment results show that Wi-Mose achieves 29.7mm and 37.8mm P-MPJPE in the LoS and NLoS scenarios, which is 21\% and 10\% increased in accuracy compared with the baseline, respectively. In the future, we will prove that Wi-Mose can also construct high-precision 3D moving human pose skeletons in different environments.

%

\begin{table}[t]   
	\setlength{\abovecaptionskip}{-0.1cm}
	\renewcommand{\arraystretch}{1.1}
	\renewcommand\tabcolsep{18.0pt}
	\caption{P-MPJPE(unit:$mm$) for The Cross-subject Scenario}
	\label{cross}
	\fontsize{8}{10}\selectfont   
	\centering
		\begin{tabular}{p{1.5cm}<{\centering}|p{1.5cm}<{\centering}|p{1.5cm}<{\centering}}
			\toprule
			
			 Scenarios & Model &  Overall  \\ \hline
			\multirow{2}{*}{ Basic}
			&  WiPose &43.7    \cr
			&  Wi-Mose &42.6   \cr\hline
			\multirow{2}{*}{ Occluded}
			&  WiPose &50.7   \cr  
			& Wi-Mose &46.8   \\
			\bottomrule
			
	\end{tabular}
	
	\vspace{-0.8em}
	
\end{table}

\ifCLASSOPTIONcaptionsoff
  \newpage
\fi

\normalem
\begin{spacing}{1}
	\bibliographystyle{IEEEtr}
	\bibliography{test}

\begin{thebibliography}{10}

\bibitem{Tool}
D.~Halperin, W.~Hu, A.~Sheth, and D.~Wetherall, ``Tool release: Gathering
  802.11n traces with channel state information,'' {\em ACM SIGCOMM Computer
  Communication Review}, vol.~41, no.~1, pp.~53--53, 2011.

\bibitem{track}
Y.~Xie, J.~Xiong, M.~Li, and K.~Jamieson, ``{mD-Track}: Leveraging
  multi-dimensionality for passive indoor {Wi-Fi} tracking,'' in {\em The 25th
  Annual International Conference on Mobile Computing and Networking},
  pp.~1--16, 2019.

\bibitem{wirol}
L.~{Guo}, X.~{Wen}, Z.~{Lu}, X.~{Shen}, and Z.~{Han}, ``{WiRoI}: Spatial region
  of interest human sensing with commodity {WiFi},'' in {\em 2019 IEEE Wireless
  Communications and Networking Conference (WCNC)}, pp.~1--6, 2019.

\bibitem{fall}
S.~Palipana, D.~Rojas, P.~Agrawal, and D.~Pesch, ``{FallDeFi}: Ubiquitous fall
  detection using commodity {Wi-Fi} devices,'' {\em Proceedings of the ACM on
  Interactive, Mobile, Wearable and Ubiquitous Technologies}, vol.~1, no.~4,
  pp.~1--25, 2018.

\bibitem{rf}
M.~Zhao, T.~Li, M.~A. Alsheikh, Y.~Tian, and D.~Katabi, ``Through-wall human
  pose estimation using radio signals,'' in {\em Computer Vision and Pattern
  Recognition (CVPR)}, 2018.

\bibitem{rf3D}
M.~Zhao, Y.~Tian, H.~Zhao, M.~A. Alsheikh, T.~Li, R.~Hristov, Z.~Kabelac,
  D.~Katabi, and A.~Torralba, ``{RF-based 3D} skeletons,'' in {\em Proceedings
  of the 2018 Conference of the ACM Special Interest Group on Data
  Communication}, pp.~267--281, 2018.

\bibitem{letters}
L.~Guo, Z.~Lu, X.~Wen, S.~Zhou, and Z.~Han, ``From signal to image: Capturing
  fine-grained human poses with commodity {Wi-Fi},'' {\em IEEE Communications
  Letters}, vol.~24, no.~4, pp.~802--806, 2019.

\bibitem{mobicom}
W.~Jiang, H.~Xue, C.~Miao, S.~Wang, and L.~Su, ``Towards {3D} human pose
  construction using {WiFi},'' in {\em MobiCom '20: The 26th Annual
  International Conference on Mobile Computing and Networking}, 2020.

\bibitem{fresnel}
D.~{Wu}, D.~{Zhang}, C.~{Xu}, H.~{Wang}, and X.~{Li}, ``Device-free {WiFi}
  human sensing: From pattern-based to model-based approaches,'' {\em IEEE
  Communications Magazine}, vol.~55, no.~10, pp.~91--97, 2017.

\bibitem{alpha}
H.-S. Fang, S.~Xie, Y.-W. Tai, and C.~Lu, ``{RMPE}: Regional multi-person pose
  estimation,'' in {\em Proceedings of the IEEE International Conference on
  Computer Vision}, pp.~2334--2343, 2017.

\bibitem{videopose}
D.~Pavllo, C.~Feichtenhofer, D.~Grangier, and M.~Auli, ``3{D} human pose
  estimation in video with temporal convolutions and semi-supervised
  training,'' in {\em Proceedings of the IEEE Conference on Computer Vision and
  Pattern Recognition}, pp.~7753--7762, 2019.

\bibitem{Adam}
D.~Kingma and J.~Ba, ``Adam: A method for stochastic optimization,'' {\em
  Computer ence}, 2014.

\bibitem{tensorflow}
M.~Abadi, P.~Barham, J.~Chen, Z.~Chen, A.~Davis, J.~Dean, M.~Devin,
  S.~Ghemawat, G.~Irving, M.~Isard, {\em et~al.}, ``Tensorflow: A system for
  large-scale machine learning,'' in {\em 12th $\{$USENIX$\}$ symposium on
  operating systems design and implementation ($\{$OSDI$\}$ 16)}, pp.~265--283,
  2016.

\end{thebibliography}
\end{spacing}

\end{document}